\documentclass[conference]{IEEEtran}
\IEEEoverridecommandlockouts

\usepackage{booktabs}
\usepackage{amsmath}
\usepackage{amssymb}
\usepackage{graphicx}  
\usepackage{float}  
\usepackage{subfigure}  
\usepackage{subfig}  
\usepackage{footnote}
\usepackage{cite}
\usepackage{amsmath,amssymb,amsfonts}
\usepackage[ruled]{algorithm2e}
\usepackage{algorithmic}
\usepackage{graphicx}
\usepackage{textcomp}
\usepackage{xcolor}
\usepackage[T1]{fontenc}
\def\BibTeX{{\rm B\kern-.05em{\sc i\kern-.025em b}\kern-.08em
    T\kern-.1667em\lower.7ex\hbox{E}\kern-.125emX}}
\begin{document}
\title{Efficient Multi-Task Reinforcement Learning via Task-Specific Action Correction*
}

\author{Jinyuan Feng$^{1,2}$\quad Min Chen$^{1}$, \quad Zhiqiang Pu$^{1,2}$ \quad Tenghai Qiu$^{1}$ \quad Jianqiang Yi$^{1,2}$\\
$^1$Institute of Automation, Chinese Academy of Sciences, China\\
$^2$University of Chinese Academy of Sciences, China\\
}

\maketitle
\pagestyle{empty}  
\thispagestyle{empty} 
\begin{abstract}
 Multi-task reinforcement learning (MTRL) demonstrate potential for enhancing the generalization of a robot, enabling it to perform multiple tasks concurrently. However, the performance of MTRL may still be susceptible to conflicts between tasks and negative interference. To facilitate efficient MTRL, we propose Task-Specific Action Correction (TSAC), a general and complementary approach designed for simultaneous learning of multiple tasks. TSAC decomposes policy learning into two separate policies: a shared policy (SP) and an action correction policy (ACP). To alleviate conflicts resulting from excessive focus on specific tasks' details in SP, ACP incorporates goal-oriented sparse rewards, enabling an agent to adopt a long-term perspective and achieve generalization across tasks. Additional rewards transform the original problem into a multi-objective MTRL problem. Furthermore, to convert the multi-objective MTRL into a single-objective formulation, TSAC assigns a virtual expected budget to the sparse rewards and employs Lagrangian method to transform a constrained single-objective optimization into an unconstrained one. Experimental evaluations conducted on Meta-World's MT10 and MT50 benchmarks demonstrate that TSAC outperforms existing state-of-the-art methods, achieving significant improvements in both sample efficiency and effective action execution.

\end{abstract}


\section{Introduction}
Empowering generalist robots through reinforcement learning is one of the essential targets of robotic learning. Reinforcement learning (RL) with the assistance of neural networks has become a crucial methodology in various domains, such as gaming\cite{silver2016mastering, silver2017mastering,berner2019dota,vinyals2019grandmaster}, large language models\cite{ouyang2022training} and real-world applications including robotics\cite{lillicrap2015continuous}. However, the majority of research in RL predominantly focuses on specific problem scenarios, prioritizing mastery of individual tasks through the learning of single policies, often at the expense of generalization. Multi-task reinforcement learning (MTRL), on the other hand, emerges as a promising approach to improve generalization by leveraging domain information obtained from related tasks.

MTRL naturally incorporates a curriculum, as it enables the learning of more manageable tasks to facilitate the teaching of more challenging tasks\cite{pinto2017learning}. However, MTRL is prone to negative transfer\cite{ruder2017overview}, a phenomenon where the task-specific knowledge from a task can impede the overall learning process of other tasks. This phenomenon is also referred to as task conflict, which becomes more acute as the number of tasks increases. From the optimization standpoint, task conflict arises from conflicting gradients\cite{yu2020gradient} between tasks, where the gradients move in opposite directions. 

A multitude of methods have been developed to alleviate negative transfer and achieve efficient MTRL. Classical approaches include policy distillation\cite{rusu2015policy,teh2017distral}, explicit measurement of task relatedness\cite{liu2022auto,fifty2021efficiently}, and information sharing\cite{yu2021conservative,yu2022leverage}, among others. Policy distillation involves training a smaller network structure to bring previous tasks to an expert level, thereby integrating multiple policies into a single policy. However, these methods increase the number of network parameters as they require separate networks for different tasks and an additional distillation step. Some researchers utilize validation loss on tasks\cite{liu2022auto} or causal influence\cite{fifty2021efficiently} to determine better task groupings. One drawback of the aforementioned methods is the need for substantial computational resources. Often, calculating task correlations and adjusting learning methods can only be done through trial and error, leading to high costs. In MTRL, information sharing can be achieved through data sharing, parameters sharing, representations sharing, or behavior sharing. For instance, CDS\cite{yu2022leverage} routes data based on task-specific data to improve information sharing. Similarly, a simple method proposed in \cite{yu2021conservative} applies a zero reward to unlabeled data, facilitating data sharing in theory and practice. Parameter sharing through learning shared representations can effectively transfer knowledge between policies. For instance, a soft modularization method presented in \cite{yang2020multi} shares parameters by generating soft combinations of different modules. Similarly, AdaShare\cite{sun2020adashare} and an automated multi-task learning algorithm in \cite{guo2020learning} adaptively determine the feature sharing mode across multiple tasks. However, these methods suffer from high computational complexity as they require dynamic exploration of network connectivity. Attention mechanisms can be leveraged to share representations, as proposed in\cite{bram2020attentive , cheng2023multi}, which group task knowledge into sub-networks without the need for prior assumptions. CARE\cite{sodhani2021multi} leverages metadata to capture the shared structure among tasks. There is another significant MTRL work that focuses on the challenge of multi-objective optimization from a gradient perspective, aiming to reduce the impact of conflicting gradients by manipulating the gradients based on various criteria\cite{yu2020gradient,liu2021conflict}. However, these methods impose an additional computational burden. The aforementioned methods reduce or coordinate conflicts between tasks from the perspectives of representation, gradient, task grouping, etc., effectively achieving MTRL. However, they neglect the potential causes of task conflict. An agent solely focuses on task-specific information and details within each individual task, it may result in short-sightedness and hinder the generalization across tasks. By introducing goal-oriented sparse learning signals, it is possible to strike a balance between task-specific performance and generalization across tasks.

\begin{figure}[h]
  \centering
  \includegraphics[trim=40 40 350 120, clip,width=0.75\linewidth]{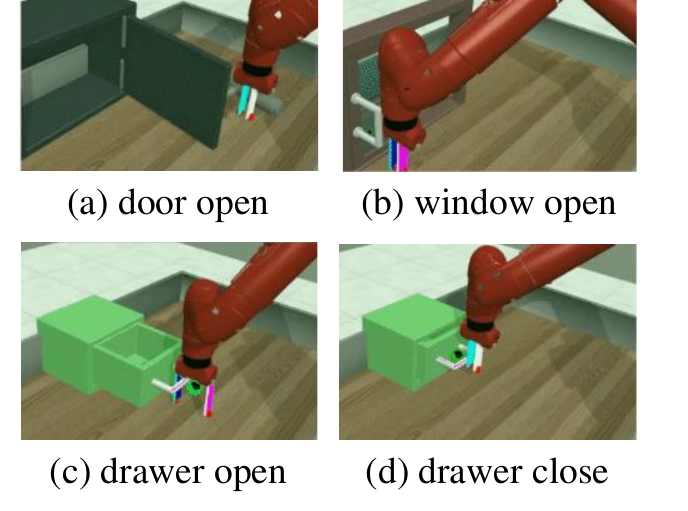}
  \caption{a variety of related manipulation tasks. Several actions are similar across these tasks: getting closer to the objects and interacting with them.}
  \label{metaworld_1}
\end{figure}

Imagine that humans are simultaneously learning a variety of related manipulation tasks, for instance, as depicted in Fig. \ref{metaworld_1}. Within these tasks, there are several actions that exhibit similarity, such as approaching objects like doors or drawers and interacting with them. Humans have the ability to leverage previously learned behaviors when encountering a specific task, making slight adjustments based on the task's goal. Furthermore, when humans are confronted with numerous tasks simultaneously, having detailed instructions and requirements for each task can lead to confusion and a sense of being overwhelmed, even though they can be helpful in completing individual tasks. In contrast, when each task has a clearly defined goal, humans naturally adopt a longer-term perspective regarding conflicts and priorities among tasks. The above human behaviors imply their recognition that current conflicts may not be necessary for achieving the overall objectives of the tasks. Our idea is inspired by this recognition.

In this paper, we propose a novel approach called \textbf{T}ask-\textbf{S}pecific \textbf{A}ction \textbf{C}orrection (TSAC) as a general and complementary method for MTRL. TSAC decomposes policy learning into two policies: a shared policy (SP) and an action correction policy (ACP). SP maximizes well-shaped and intensive rewards, which focus on task-specific information and accelerate the learning process. Its output actions, referred to as shared actions, are potentially short-sighted. On the other hand, ACP utilizes goal-oriented sparse rewards to generate a far-sighted edited action that can cross tasks. The goal-oriented sparse rewards which is sparse and strongly correlated with the completion of the objective. SP and ACP collaborate with each other, where SP provides a suboptimal policy that facilitates the training of ACP in the sparse rewards setting. ACP, in return, improves the overall performance. To balance the training of these two policies, we assign a virtual expected budget to the sparse rewards and use the Lagrangian method to dynamically adjust the weights of the loss in ACP. Our two-policy paradigm draws inspiration from works in safe reinforcement learning\cite{yu2022towards,zhang2020first,qin2021density}. However, our approach differs significantly in terms of motivation and interpretation.

To implement our approach, we employ the Soft Actor-Critic algorithm\cite{haarnoja2018soft} as the underlying reinforcement learning policy. It is worth noting that our approach is algorithm-agnostic and can be integrated with existing MTRL methods. Our experimental results demonstrate the efficiency and significance of the cooperation between the two policies and simply combining the two rewards do not yield comparable results. Moreover, our experiments conducted on the Metaworld benchmark\cite{yu2020meta} using MT10 and MT50 showcase significant improvements in both sample efficiency and final performance compared to previous state-of-the-art multi-task policies. In summary, our contributions are as follows:

(1) We propose {T}ask-\textbf{S}pecific \textbf{A}ction \textbf{C}orrection (TSAC), a general and complementary approach for MTRL, which decomposes policy learning into two policies, facilitating efficient MTRL. TSAC can be combined with any existing MTRL methods.

(2) We introduce goal-oriented sparse rewards to provide agents with a long-term perspective for handling task conflicts that arise from excessive focus on specific tasks' details.

(3) We assign a virtual expected budget to the sparse rewards and utilize Lagrangian method to transform a constrained single-objective optimization into an unconstrained one. The Lagrangian multiplier dynamically adjust the loss weights in the two policy networks.

(4) We demonstrate the efficiency and significance of the cooperation between the two policies in MTRL, and show that TSAC achieves significant improvements in both sample efficiency and effective action execution.

\section{preliminaries}
\subsection{Multi-Task Reinforcement Learning}
We extend the single-task Markov Decision Process (MDP) problem to multi-task MDP for a agent under the framework of Contextual MDP (CMDP)\cite{hallak2015contextual}. CMDP is defined by a tuple $\left\langle\mathcal{C}, \mathcal{S}, \mathcal{A}, \gamma, \mathcal{M}\right\rangle$. Here, $\mathcal{C}$ can be viewd as a task set $\mathcal{C}=\left\{\mathcal{T}_i\right\}_{i=1}^N$, where $\mathcal{T}_i$ denotes the task $i$ and $N$ is the number of tasks. $\mathcal{S}$ represents the shared state space , $\mathcal{A}$ denotes the shared action space, and $\gamma$ is the discount factor. State $s \in \mathcal{S}$ and action $a \in \mathcal{A}$. $\mathcal{M}$ is a fuction that maps a task $\mathcal{T}_i \in \mathcal{C}$ to MDP parameters, such that $\mathcal{M}(\mathcal{T}_i)=\left\{P_{i},  R_{i}, \rho_i\right\}$. The transition probability $P_{i}$, reward function $R_{i}$ and initial state distribution $\rho_i$ vary by each task. During training, tasks are sampled from a uniform distribution $p(\mathcal{T})$. The agent's policy takes state $s$ as input and outputs action $a$. The objective of the agent's policy $\pi$ is to maximize the expected return $\mathbb{E}_{\mathcal{T}_i\sim p(\mathcal{T})}[\mathbb{E}_\pi[\sum_t\gamma^tR_i(s_t,a_t)]]$, where $s_t$ and $a_t$ represent the state and the action at timestep $t$.

\subsection{Soft Actor-Critic}
In this paper, we adopt Soft Actor-Critic (SAC) \cite{haarnoja2018soft} as the fundamental policy. As observed in \cite{yu2020meta}, SAC, being an off-policy actor-critic algorithm with a strong exploration ability based on maximum entropy, exhibits superior performance. 

The concept of SAC goes beyond merely maximizing cumulative rewards. It also introduces additional stochasticity to the policy. Thus, a regularization term that incorporates entropy is included in the reinforcement learning objective. The correction term added with entropy can be defined as follows:
\begin{equation}
  \pi^{*}=\arg \max _{\pi} \mathbb{E}_{\pi}\left[\sum_{t} r\left(s_{t}, a_{t}\right)+\alpha H\left(\pi\left(\cdot \mid s_{t}\right)\right)\right]    
\end{equation}
In this context, $\alpha$ represents a regularization coefficient that controls the significance of the entropy term. $H$ denotes the entropy.

\section{method}
In this section, we present a general and complementary approach named \textbf{T}ask-\textbf{S}pecific \textbf{A}ction \textbf{C}orrection (TSAC) to decompose policy learning across two policies: the shared policy (SP) and the action correction policy (ACP).

\subsection{Overall Structure of TSAC}
\begin{figure}[h]
  \centering
  \includegraphics[trim=200 20 150 30, clip,width=0.8\linewidth]{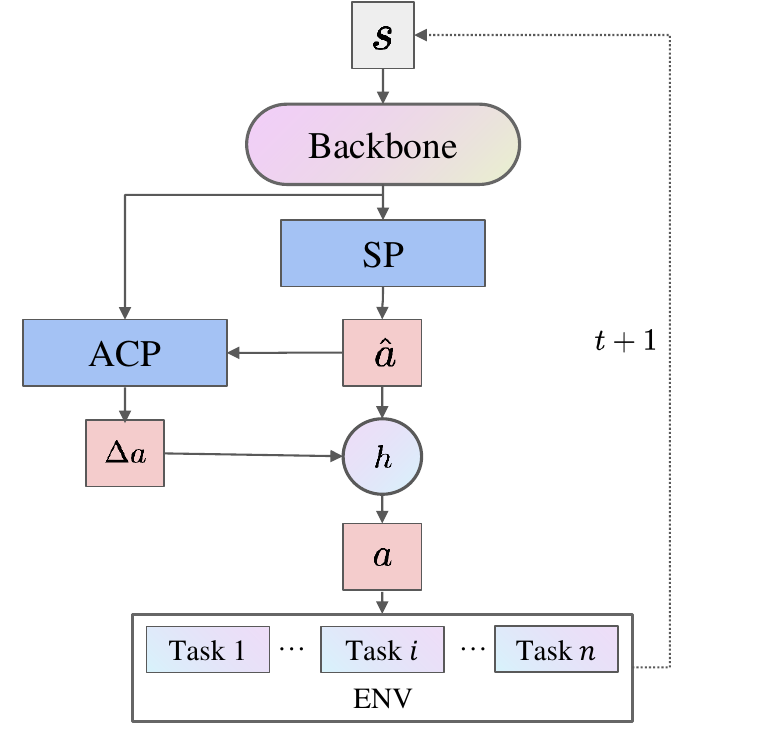}
  \caption{The structure of TSAC with two policies: a shared policy (SP) and an action correction policy (ACP).}
  \label{framework}
\end{figure}

As illustrated in Fig. \ref{framework}, TSAC consists of a pair of cooperative policies. The first policy, called the shared policy (SP) and denoted as $\operatorname{\pi}_\phi$, optimizes the guiding dense rewards by proposing a preliminary action $\hat{a}\sim\pi_{\phi}(\cdot|s)$. However, this preliminary action may be shortsighted. The second policy, referred to as the action correction policy (ACP) and denoted as $\pi_\psi$, corrects the preliminary action by providing an action correction $\Delta a\sim\pi_{\psi}(\cdot|s,\hat{a})$ to execute an effective action. The result action, denoted as $a=h(\hat{a},\Delta a)$, is then output to the environment, where $h$ represents an editing function. ACP is conditioned on SP's output $\hat{a}$. Together, these two policies cooperate to improve sample efficiency and performance. For simplicity, we denote the overall composed policy as $\pi_{\psi\circ\phi}(a|s)$.

\textbf{Motivation:} TSAC decomposes policy learning into two subtasks that focus on guiding dense rewards or goal-oriented sparse rewards. This decomposition is motivated by the following considerations:
\subsubsection{Different Conflict Horizons}
To achieve effective MTRL, most applications incorporate guiding dense rewards into each task. However, SP is myopic as it only focus on specific tasks' details and overlooks whether the final goal is accomplished, despite the final goal being the most important aspect. SP only coordinates task conflicts from various tasks in the short term. In contrast, ACP aims to maximize goal-oriented sparse rewards, enabling an agent to adopt a long-term perspective and achieve generalization across tasks. With the assistance of SP, ACP has more opportunities to obtain goal-oriented sparse rewards, thus alleviating the challenge of learning sparse rewards in ACP.
\subsubsection{Efficient Exploration}
From the perspective of SP, its action is altered by ACP. Instead of discouraging SP from taking suboptimal actions, ACP offers suggestions to improve the preliminary action, enabling SP to continue exploration in an effective and far-sighted manner. This guarded exploration leads to a better overall exploration policy because ACP is less likely to hinder SP’s exploration. From the perspective of ACP, it determines the action based on SP, enhancing its ability to explore. From an entropy perspective, the decomposition of policy learning into two policy networks introduces additional entropy.

\subsection{Goal-oriented sparse rewards}
Manually designed dense rewards incorporate prior knowledge and effectively guide policy learning. However, the magnitude of reward values does not directly indicate the ability of a policy to accomplish tasks. Therefore, we have introduced goal-oriented sparse rewards, which are correlated with the completion of the objective. The goal-oriented sparse rewards of a task $\mathcal{T}_i$ is characterized by an "$\epsilon$-region" in state space, represented by:
\begin{equation}
    R_i^s(s,a)=\begin{cases}\delta_{s_g}(s)&\mathrm{if~}f(s,{s}_g)\leq\epsilon\\0&\mathrm{else},\end{cases}
\end{equation}
In this equation, $R_i^s$ denotes the goal-oriented sparse rewards of task $\mathcal{T}_i$, $s$ is the current state, $s_g$ denotes the goal state, $f(s,s_g)$ is a function that maps the goal state and current state to a latent space, computing the distance between them. $\delta_{s_g}$ defines the reward value and set $\delta_{s_g}=1$. $\epsilon$ represents a small distance threshold. 

Following the description of CMDP (section \uppercase\expandafter{\romannumeral2}. A), the initial state distribution $\rho_i$ determines the probability density of an episode starting at state $s_{0}$. For each transition $(s_{t},a_{t},s_{t+1})$ from task $\mathcal{T}_i$ at timestep $t$, the environment produces a scalar $R_i(s_t,a_t)$. It is worth mentioning that the reward function $R_{i}$ is artificially designed and intensive, which we refer to as guiding dense rewards. Similarly, the environment produces a scalar $ R_i^s(s_t,a_t) $ as a goal-oriented sparse reward. Higher values for both the guiding dense rewards and the goal-oriented sparse rewards indicate better performance.

For each state $s_t$, the guiding dense rewards state value of policy $\pi$ is denoted as $V^{\pi}(s_t)=\mathbb{E}_{\mathcal{T}_i\sim p(\mathcal{T})}[\mathbb{E}_{\pi}\sum_{t^{\prime}=t}^{\infty}\gamma^{t^{\prime}-t}R_i(s_{t^{\prime}},a_{t^{\prime}})]$. The guiding dense rewards state-action value is denoted as $Q^{\pi}(s_{t},a_{t})=\mathbb{E}_{\mathcal{T}_i\sim p(\mathcal{T})}[R_i(s_{t},a_{t})+\gamma\mathbb{E}_{s_{t+1}\thicksim P_i}V^{\pi}(s_{t+1})]$. Similarly, We define $V_s^{\pi}$ and $Q_s^{\pi}$ for the goal-oriented sparse rewards. 

We consider the MTRL objective from the perspective of guiding dense rewards and goal-oriented sparse rewards:
\begin{equation}
    \{\max_{\pi}\mathbb{E}_{\mathcal{T}_i\sim p(\mathcal{T})}\underset{s_0\sim\rho_i}{\operatorname*{\mathbb{E}}}V^{\pi}(s_0),
    \max_{\pi}\mathbb{E}_{\mathcal{T}_i\sim p(\mathcal{T})}\underset{s_0\sim\rho_i}{\operatorname*{\mathbb{E}}}V_s^{\pi}(s_0)\}
\end{equation}
MTRL can be viewed as a multi-objective optimization problem. The introduction of sparse rewards adds an additional objective to the MTRL framework, which increases the problem's complexity. To convert multi-objective MTRL into single-objective MTRL, we assign a virtual expected budget $C$ to the sparse rewards. This allows us to write the MTRL objective as follows:
\begin{equation}
\begin{split}
    \max_\pi\mathbb{E}_{\mathcal{T}_i\sim p(\mathcal{T})}\underset{s_0\sim\rho_i}{\operatorname*{\mathbb{E}}}V^\pi(s_0), \\
    s.t.\quad \mathbb{E}_{\mathcal{T}_i\sim p(\mathcal{T})}\underset{s_0\sim\rho_i}{\operatorname*{\mathbb{E}}}V_s^\pi(s_0)+C\geq0,    
    \label{constrained eq}
\end{split}
\end{equation}
In this equation, $C$ represents a virtual expected budget. Specifically for each time step, Eq.\ref{constrained eq} can be rewritten as:
\begin{equation}
    \mathbb{E}_{\mathcal{T}_i\sim p(\mathcal{T})}[\mathbb{E}_\pi[\sum_t\gamma^t(R_i^s(s_t,a_t)+c)]]\geq0,
    \label{exp_budget}
\end{equation}
Here, $c$ denotes the expected budget specific to each step and it relates to $C$ through the equation $\sum_{t=0}^\infty\gamma^tc=\frac c{1-\gamma}=C$. Notably, the expected budget represents an average target to be achieved rather than a strict enforcement.

To simplify the problem, we transform the multi-objective MTRL into a single-objective MTRL using the Lagrangian method. This method converts the constrained optimization problem Eq.\eqref{constrained eq} into an unconstrained one by introducing a multiplier $\lambda$:
\begin{equation}
\begin{split}
        &\min_{\lambda\geq0}\max_{\pi}\mathbb{E}_{\mathcal{T}_i\sim p(\mathcal{T})}\underset{s_0\sim\rho_i}{\operatorname*{\mathbb{E}}}V^\pi(s_0)\\
    &+\lambda\left(\mathbb{E}_{\mathcal{T}_i\sim p(\mathcal{T})}\underset{s_0\sim\rho_i}{\operatorname*{\mathbb{E}}}V_s^\pi(s_0)+C\right).
    \label{optimize_object}
\end{split}
\end{equation}
In this formulation, the weight of the goal-oriented sparse rewards combined with the guiding dense rewards is represented by $\lambda$. We solve this objective by using model-free MTRL algorithms.

\subsection{Objective function}
We utilize an off-policy actor-critic approach to train the two policies. Given the overall policy $\pi_{\psi\circ\phi}(a|s)$, we use the typical Temporal Difference (TD)\cite{tesauro1995temporal} backup to learn $Q^{\pi_{\psi\circ\phi}}(s,a;\theta)$ and $Q_s^{\pi_{\psi\circ\phi}}(s,a;\theta_s)$, which are parameterized as $Q(s,a;\theta)$ and $Q_s(s,a;\theta_s)$ respectively. Here, $\theta$ represents the network parameters collectively for the two state-action values. When provided with $s_{t+1}$ and $a_{t+1}$, the Bellman backup operator for the guiding dense rewards state-action value is expressed as:
\begin{equation}
   Q(s_t,a_t;\theta)=\mathbb{E}_{\mathcal{T}_i\sim p(\mathcal{T})}R_i(s_t,a_t)+\gamma Q(s_{t+1},a_{t+1};\theta),
   \label{critic}
\end{equation}
where $R_i$ represents the guiding dense rewards obtained from task $\mathcal{T}_i$. The backup operator for the goal-oriented sparse rewards state-action value $Q_s$ is defined similarly with $R^s$. Both $Q(s,a;\theta)$ and $Q_s(s,a;\theta_s)$ can be learned from transitions $(s_t,a_t,s_{t+1})$ sampled from a replay buffer.

To achieve off-policy training of $\psi$ and $\phi$, we convert Eq.\eqref{optimize_object} into a bi-level optimization surrogate. The formulation is presented as follows:
\begin{equation}
\begin{split}
        &\mathrm{(a)}\quad\max_{\phi,\psi}\underset{s\sim\mathcal{D}}{\operatorname*{\mathbb{E}}}\left[Q(s,a;\theta)+\lambda Q_s(s,a;\theta_s))\right],
        \\
        &\mathrm{(b)}\quad\min_{\lambda\geq0}\lambda\Lambda_{\pi_{\psi\circ\phi}}.\quad \Lambda_{\pi_{\psi\circ\phi}}\triangleq\underset{s_0\sim\rho_i}{\operatorname*{\mathbb{E}}}V_s^{\pi_{\psi\circ\phi}}(s_0)+C
\end{split}
\label{bi-level optimization}
\end{equation}
Here, $\mathcal{D}$ represents a replay buffer containing a historical marginal state distribution used to train the policies. The initial state distribution $\rho$ is employed to train $\lambda$. This distinction arises from the idea that when fine-tuning $\lambda$, we should primarily consider how well the policy satisfies the virtual expected budget starting with $\rho$, rather than with some historical state distribution. Subsequently, We further transform the off-policy objective(Eq.\eqref{bi-level optimization},a) into two parts:
\begin{equation}
\begin{split}
        &\mathrm{(a)}\quad\max_{\textcolor{blue}{\phi}}\underset{\substack{s \sim \mathcal{D}, \textcolor{blue}{\hat{a}} \sim \pi_{\textcolor{blue}{\phi}}(\cdot \mid s),\\ \textcolor{blue}{\Delta a} \sim \pi_{\psi}(\cdot \mid s, \textcolor{blue}{\hat{a}}) \\ \textcolor{blue}{a}=h(\textcolor{blue}{\hat{a}}, \textcolor{blue}{\Delta a})}}{\mathbb{E}}[Q(s, \textcolor{blue}{a};\theta],
        \\
        &\mathrm{(b)}\quad\max_{\textcolor{red}{\phi}}\underset{\substack{s \sim \mathcal{D}, \hat{a} \sim \\pi_{\textcolor{blue}{\phi}}(\cdot \mid s),\\ \textcolor{red}{\Delta a} \sim \pi_{\textcolor{red}{\psi}}(\cdot \mid s, \textcolor{blue}{\hat{a}}) \\ \textcolor{red}{a}=h(\hat{a}, \textcolor{red}{\Delta a})}}{\mathbb{E}}[-d(\textcolor{red}{a},\hat{a})+ \lambda Q_s(s, \textcolor{red}{a};{\theta}_s)],
\end{split}
\label{bi-level transform}
\end{equation}
In this modified formulation, the distance function $d(a,\hat{a})$ quantifies the change from $\hat{a}$ to $a$. It is not necessarily proportional to $\Delta a$ as the editing function $h$ can introduce nonlinearity. ACP $\pi_{\psi}$ aims to maximize the goal-oriented sparse rewards while minimizing the distance between the actions before and after the correction. On the other hand, SP $\pi_{\phi}$ solely focus on maximizing the guiding dense rewards. Importantly, ACP modifies SP's action, which aligns with the discussed motivation for efficient exploration. The training objective (Eq.\eqref{bi-level transform},b) for ACP relies on a critic $Q_s$, which learns the expected future goal-oriented sparse rewards. Consequently, guided by $Q_s$, ACP explores actions with greater potential in long-term sequences.

\subsection{Action correction function and Distance function}
In this section, we present the design for the action correction function $h(\hat{a},\Delta a)$ and the distance function $d(a,\hat{a})$.
\subsubsection{Action correction function}
We opt for the correction function $h$ to be primarily additive and simple, which helps to reduce training difficulty. Without loss of generality, we assume a bounded action space $[-A,A]$, and that both $\hat{a}$ and $\Delta a$ are already within this space. Consequently, we define:
\begin{equation}
   a=h(\hat{a},\Delta a)=\operatorname*{min}(\operatorname*{max}(2\hat{a}+\Delta a,-A),A),  
\end{equation}
Here, the element-wise min and max, along with the multiplication by $2$ and the clipping, ensure that $a\in[-A,A]$. This means that SP retains full control over the final action and can overwrite the action if necessary. This is crucial because ACP faces challenges in learning an effective correction policy in the short horizon when SP still dominates the learning process. Although the additive operation is simple, the overall editing process is sufficiently general to encompass any modification.

This additive editing function is motivated by the goal of achieving sparsity. Policies are typically evaluated based on metrics such as success, which are only triggered for certain states. To explicitly incorporate this inductive bias, we adopt the additive correction function, which ensures that ACP learns a policy that maximizes metrics such as success based on SP. This simplifies the optimization landscape of ACP.

\subsubsection{Distance function}
we utilize the hinge loss to compare the guiding dense rewards state-action values of $\hat{a}$ and $a$:
\begin{equation}
    d(a,\hat{a})\triangleq\max(0,Q(s,\hat{a};\theta)-Q(s,a;\theta))
\end{equation}
Here, $Q$ represents the critic and $\theta$ denotes the parameters. This loss yields zero if the edited action $a$ already attains a higher state-action value than the preliminary action $\hat{a}$. In such cases, only $Q_s$ is optimized by $\pi_{\psi}$. Otherwise, the inner part of (Eq.\eqref{bi-level transform},b) is recovered as $Q(s,a)+ \lambda Q_s(s,a)$. Our distance function in the critic $Q$ is more appropriate than $L_2$ distance in the action space, because we ultimately care about how the $Q$ changes after the action is edited.

\subsection{Training process}
To practically train the objectives, we employ stochastic gradient descent (SGD) simultaneously to Eq.\eqref{bi-level optimization}(b) and Eq.\eqref{bi-level transform}. The re-parameterization trick is utilized for both $\pi_{\psi}$ and $\pi_{\phi}$ to enable the application of SGD. To assess $\Lambda_{\pi_{\psi\circ\phi}}$, a batch of rollout experiences $\{(s_n,a_n)\}_{n=1}^N$ following $\pi_{\psi\circ\phi}$ is provided. The gradient of $\lambda$ (Eq.\eqref{bi-level optimization}(b)) is approximated as:
\begin{equation}
    \Lambda_{\pi_{\psi \circ \phi}} \approx \frac{1}{N} \sum_{n=1}^{N} R_{s}\left(s_{n}, a_{n}\right)+c
    \label{estimatelambda}
\end{equation}
Here, $c$ represents the virtual expected budget as defined in Eq.\eqref{exp_budget}. Subsequent to each rollout, a batch of goal-oriented sparse rewards is collected, each reward is compared to $-c$, and the mean of the differences is used to adjust $\lambda$. This approximation enables the updating of $\lambda$ using mini-batches of data, irather than waiting for complete episodes to conclude or relying on the often inaccurate estimated $V_s^{\pi\psi\circ\phi}$. Multiple parallel environments are employed to mitigate temporal correlation within the rollout batch data, which constitutes the data to be placed into the replay buffer.
\begin{figure}[h]
  \centering
  \includegraphics[trim=200 70 200 50, clip,width=0.8\linewidth]{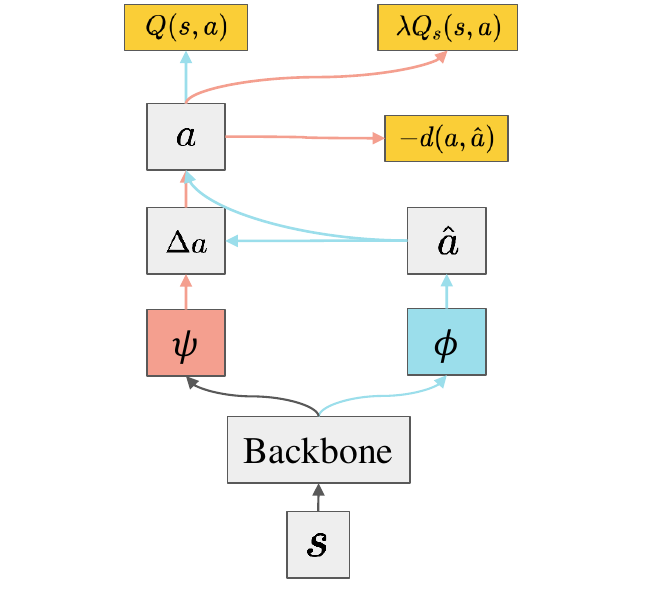}
  \caption{The computation graph of Eq.\eqref{bi-level transform}. Nodes denote variables or networks and edges denote operations. The orange blocks are negative losses, the blue paths are the gradient paths of $\phi$, and the red paths are the gradient paths of $\psi$.}
  \label{computation_graph}
\end{figure}

The computational graph presenting Eq.\eqref{bi-level transform} is illustrated in Fig. \ref{computation_graph}. Our approach is applicable to various goal-oriented sparse rewards and action distance functions. To encourage exploration, we integrate SAC\cite{haarnoja2018soft}, which incorporates the entropy terms of $\pi_{\psi}$ and $\pi_{\phi}$ in Eq.\eqref{bi-level transform}, dynamically adjusting their weights based on two entropy targets as described in \cite{haarnoja2018soft}.In our experiments, both SP and ACP are trained from scratch. The pseudocode for the overall TSAC is shown in the algorithm 1.

\begin{algorithm}[h]  
	\caption{TSAC: Task-Specific Action Correction}
	\LinesNumbered 
	\KwIn{$N$ tasks; virtual expected budget $C$}
	  Initialize $\theta$,$\phi$ and $\psi$; reset the replay buffer $\mathcal{D}\leftarrow\emptyset$ \\  
	\For{each training iteration}{
		Reset the rollout batch $\mathcal{B}\leftarrow\emptyset $\;
		\For{each rollout step}{
		    \For{task $i \cdot\cdot\cdot N$}{
		    	Action proposal by SP: $\hat{a}\sim\pi_{\phi}(\cdot|s)$\;
		        Action correcting by ACP: $\Delta a\sim\pi_{\psi}(\cdot|s,\hat{a})$\;
		        Output action $h(\hat{a},\Delta a)$\;
		        Task $i$'s transition $s^{\prime}\sim\mathcal{P}_i(s^{\prime}|s,a)$\;
		        Add the transition to the rollout batch $\mathcal{B}\leftarrow\mathcal{B}\bigcup\{(s,a,s^{\prime},R_i(s,a),R_i^s(s,a))\}$\;
		    }
		}
		Store the rollout batch in the buffer $\mathcal{D}\leftarrow\mathcal{D}\bigcup\mathcal{B}$\;
		Sample a training mini-batch $\mathcal{B}_t$ from the replay buffer $\mathcal{D}$ for computing gradient\;
		Perform one gradient step on the critic parameters $\theta$ by TD backup (Eq.\ref{critic}) on $Q$ and $Q_c$\;
		Estimate the gradient of the Lagrangian multiplier $\lambda$ by evaluating Eq.\ref{estimatelambda} on $\mathcal{B}_t$\;
		Optimize the multiplier by $\lambda \leftarrow \lambda-\alpha \Lambda_{\pi_{\psi \circ \phi}}$\;
		Use SGD to optimize SP: $\phi \leftarrow \phi+\alpha \Delta \phi$ (gradient of Eq.\ref{bi-level transform}, a)\;
		Use SGD to optimize ACP: $\psi \leftarrow \psi+\alpha \Delta \psi$ (gradient of Eq.\ref{bi-level transform}, b)\;
		Update other parameters such as entropy weight, target critic network, etc.
	}
\end{algorithm}

\section{Experiments}
\begin{figure}[h]
  \centering
  \includegraphics[width=\linewidth]{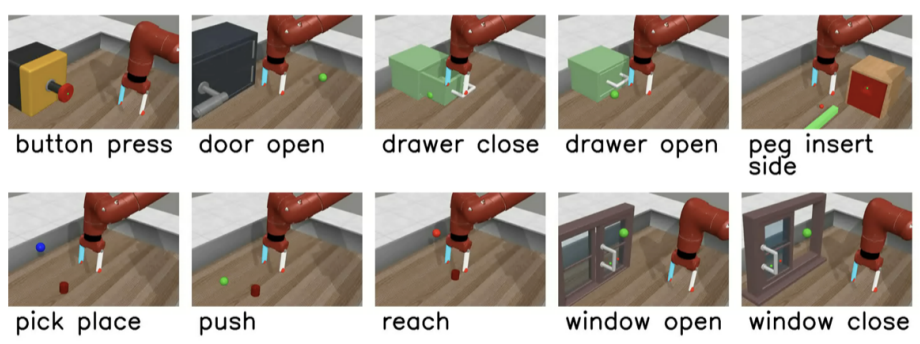}
  \caption{The MT10 benchmark from Meta-World contains 10 tasks: reach, push, pick, open window and so on.}
  \label{MT10}
\end{figure}

In this section, we evaluate TSAC in the Meta-World multi-task RL environment\cite{yu2020meta} and use Meta-World's MT10 and MT50 benchmarks. The MT10 and MT50 evaluation protocols consist of 10 and 50 tasks, respectively (shown in Fig. \ref{MT10}). We compare TSAC against several baseline methods and conduct ablation studies to verify the effectiveness of our method.

The first goal of our experimental evaluation is to assess Whether TSAC improves the performance of a multi-task agent. We compare the performance of TSAC in two different settings: a short horizon to evaluate its sample efficiency and a long horizon to measure its overall performance. For comparison, We select CARE\cite{sodhani2021multi}, MT--SAC, Soft Modularization\cite{yang2020multi} and PCGrad\cite{yu2020gradient} as our baselines. 

Furthermore, we evaluate different action correction functions to identify which yields the best performance. The action correction functions considered are SP-dominated, ACP-dominated, equal, and Softclip (See Section \uppercase\expandafter{\romannumeral4}-C for the definitions) .
\subsection{Baselines}
 We will compare our method against the following baselines:
 
 \textbf{CARE:} As a representations sharing method, representations are shared to learn how to compose them by leveraging additional information about each task.
 
 \textbf{MT--SAC:} This approach directly applies the SAC algorithm to the multi-task setting. It utilizes a shared backbone with disentangled alphas.

 \textbf{Soft Modularization:} As a parameters sharing method, Soft Modularization shares parameters and uses a routing network to softly combine all possible routes for each task.
 
 \textbf{PCGrad:} It is a gradient manipulation method that projects a task's gradient onto the normal plane of any other conflicting task. However, it has high time complexity and is not suitable for MT50 in the long horizon.
 
 \textbf{TSAC(ours):} Our proposed method builds upon CARE, leveraging its representation-sharing module. In contrast, our approach is based on behavior sharing and utilizes goal-oriented sparse rewards.
 
\begin{figure*}[h]
  \centering
  \includegraphics[trim=55 200 65 210, clip,width=\linewidth]{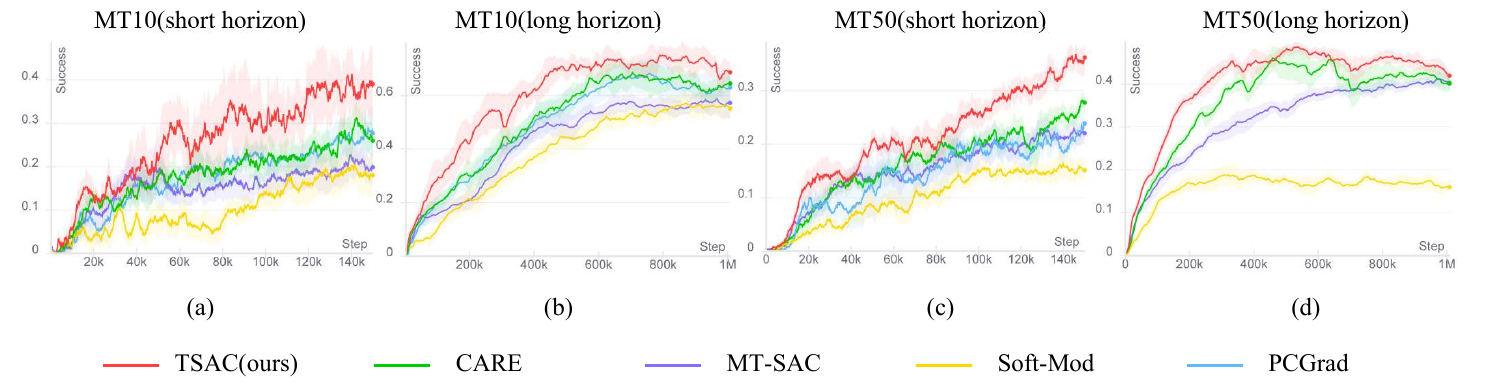}
  \caption{Training curves of different methods on all benchmarks. The bolded lines represents the mean over 4 runs for both the short horizon and long horizon. The shaded area represents the standard error.}
  \label{expriment1}
\end{figure*}

\begin{table}[!htbp]
\caption{Success rate of baselines on the short horizon(150k steps per task) for MT10 and MT50.}
\centering
\begin{tabular}{cccc}
\toprule[2pt] 
Success——150K & MT10 & MT50 \\
\midrule 
\textbf{TSAC(ours)} & \textbf{0.390 $\pm$ 0.115} & \textbf{0.362 $\pm$ 0.051}  \\
CARE & 0.260 $\pm$ 0.062 & 0.277 $\pm$ 0.028  \\
MT-SAC & 0.198 $\pm$ 0.068 & 0.220 $\pm$ 0.035 \\
Soft-Mod& 0.180 $\pm$ 0.108 & 0.151 $\pm$ 0.040 \\
PCGrad& 0.276 $\pm$ 0.107 & 0.238 $\pm$ 0.035 \\
\bottomrule[2pt] 
\label{table1}
\end{tabular}
\end{table}

\begin{table}[!htbp]
\caption{Success rate of baselines on MT10 on the long-horizon(1M steps per task). Results are reported at the end of the 0.8M,1M steps and at the best average value.}
\centering
\begin{tabular}{cccc}
\toprule[2pt] 
Success——MT10 & 0.8M & 1M & Best \\
\midrule 
\textbf{TSAC(ours)} & \textbf{0.762 $\pm$ 0.109} & \textbf{0.722 $\pm$ 0.052} & \textbf{0.827 $\pm$ 0.038} \\
CARE & 0.667 $\pm$ 0.082 & 0.642 $\pm$ 0.076 & 0.708 $\pm$ 0.114 \\
MT-SAC & 0.574 $\pm$ 0.097 & 0.555 $\pm$ 0.166 & 0.635 $\pm$ 0.120 \\
Soft Mod& 0.549 $\pm$ 0.087 & 0.560 $\pm$ 0.107 & 0.596 $\pm$ 0.102 \\
PCGrad& 0.655 $\pm$ 0.150 & 0.644 $\pm$ 0.090 & 0.708 $\pm$ 0.114 \\
\bottomrule[2pt] 
\label{table2}
\end{tabular}
\end{table}

\begin{table}[!htbp]
\caption{Success rate of baselines on MT50 on the long-horizon(1M steps per task). Results are reported at the end of the 0.8M,1M steps and at the best average value.}
\centering
\begin{tabular}{cccc}
\toprule[2pt] 
Success——MT50 & 0.8M & 1M & Best \\
\midrule 
\textbf{TSAC(ours)} & \textbf{0.450 $\pm$ 0.046} & \textbf{0.445 $\pm$ 0.045} & \textbf{0.524 $\pm$ 0.030} \\
CARE & 0.418 $\pm$ 0.057 & 0.395 $\pm$ 0.031 & 0.497 $\pm$ 0.035 \\
MT-SAC & 0.381 $\pm$ 0.044 & 0.390 $\pm$ 0.045 & 0.431 $\pm$ 0.046 \\
Soft Mod& 0.155 $\pm$ 0.033 & 0.162 $\pm$ 0.034 & 0.207 $\pm$ 0.051 \\
\bottomrule[2pt] 
\label{table3}
\end{tabular}
\end{table}

\subsection{Comparative evaluation}
Fig. \ref{expriment1}a shows the average success rate on the 10 tasks of the MT10 benchmark form Meta-world for TSAC, CARE, MT--SAC, Soft Modularization, and PCGrad. Since the success rate is a binary variable, it is noisy; therefore, the results were averaged across multiple seeds and the curves were smoothed. Mean and standard error are reported for each value.

We consider 1 million steps as a long horizon, and 150 thousand steps as a short horizon. The short horizon is utilized to observe the exploration ability of different methods, while the long horizon is used to visualize the performance of the method at various time points. It is worth noting that all methods are trained using SAC with disentangled alphas.

Table \ref{table1} and Fig. \ref{expriment1}a demonstrate that our method outperforms all the baselines on the short horizon in MT10. Additionally, Fig. \ref{expriment1}c illustrates that even in MT50 our method still outperform all baselines. For comparison, it takes around 1 million steps for the Multi-task SAC agent to reach the accuracy that our TSAC agent achieves around 300 thousand steps, suggesting that our method is highly sample-efficient and exploration-efficient.

Table \ref{table2} and \ref{table3} as well as Fig. \ref{expriment1}b and Fig. \ref{expriment1}d depict that TSAC is able to learn a good policy on the long horizon. For MT10, TSAC performs best and reaches a top success rate of 0.827 during the training. Furthermore, sampling the success rate around 0.8 million and 1 million steps reveals that TSAC has the best performance. For MT50, conflicts between tasks become more acute. CARE stops learning around 600 thousand steps and performance begins to decline in the following training steps. Despite suffering from the conflicts between tasks, TSAC's performance declines, but it still performs better than CARE. MT-SAC achieves smooth learning and has similar performance to that of CARE and TSAC at 0.8 million and 1 million steps. However, in terms of the best performance throughout training, MT--SAC is far inferior to TSAC, which outperforms all methods. Since the success metric is a binary variable and very noisy, the best performance is obtained by smoothing over the curve. Importantly, TSAC achieves average success rates of 0.450 and 0.445 on MT50 at 0.8 million and 1 million steps, surpassing the reported results from CARE, MT--SAC and Soft Modularization.

\subsection{Ablation study}
The result in the previous section suggests that TSAC is both sample efficient and exploration efficient. Furthermore, TSAC yields a good policy on long horizons. In this section, we further investigate the impact of the various action correction functions employed by TSAC and discuss the potential for improved performance.

 In ablation study, We will consider four different functions $h$:
 
 \textbf{SP dominated (ours)}: $h=\operatorname*{min}(\operatorname*{max}(2\hat{a}+\Delta a,-A),A)$, SP's action dominates the final action.
 
 \textbf{ACP dominated}: $h=\operatorname*{min}(\operatorname*{max}(\hat{a}+2\Delta a,-A),A)$, ACP's action dominates the final action.
  
 \textbf{equal contribution}: $h=\operatorname*{min}(\operatorname*{max}(\hat{a}+\Delta a,-A),A)$, SP and ACP contribute equally to the final action.
   
 \textbf{Softclip}: $h=Softclip(2\hat{a}+\Delta a)$, $h$ use softclip to smooth out the output action and bring in nonlinearity.
 
\begin{figure}[h]
  \centering
  \includegraphics[trim=230 130 280 80, clip,width=0.8\linewidth]{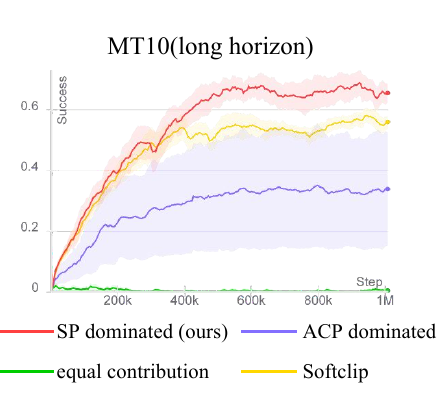}
  \caption{Training curves of different action correction function on MT10. The bolded line represents the mean over 4 runs. The shaded area represents the standard error. }
  \label{ablation}
\end{figure}

Fig. \ref{ablation} illustrates that SP dominated action correction function outperforms other functions in producing the final action. However, when SP and ACP contribute equally to the final action, the agent encounters diffculties in learning a policy. This conflict relationship between SP and ACP presents challenges in optimizing each respective goal. Conversely, when either SP or ACP dominates the action correction funcion, the agent can successfully learn a well-performing policy. Comparatively, SP outperforms ACP due to ACP's focus on optimizing the goal-oriented sparse rewards, which is challenging to train due to its sparsity. In contrast, the Softclip function exhibits a slight decrease in performance compared to our action correction function. This decline can be attributed to the introduction of nonlinearity, which further complicates and hampers the learning process. Consequently, we opt for a simple and linear operation instead of training a learnable network.

\section{CONCLUSION AND FURTHER WORK}
In this work, we present the \textbf{Task-Specific Action Correction (TSAC)}, a general and complementary MTRL method inspired by Safe Reinforcement Learning, that surpasses the several well-performed baselines on the MT10 and MT50 benchmark from Meta-World.

In this paper, we showed that our method is able to learn a high-performing policy and achieve significant improvements in both sample efficiency and final performance compared to previous state-of-the-art multi-task policies. Furthermore, TSAC is general and complementary enough to be integrated with existing methods like CARE, MT--SAC to improve them. Finally, we show the benifits of decomposing policy learning into two policies and the validity of introducing goal-oriented sparse rewards. TSAC still performs well with more difficult and diverse tasks (MT50).

In future work, We will introduce a pre-trained paradigm which policy is pretrained to maximize guiding dense rewards, and this policy is used as the initialization for SP. In this case, SP cannot be frozen because ACP continuously modifies its MDP. Instead, SP needs to be fine-tuned to adapt to the evolving actions of ACP. This pre-trained SP has the potential to accelerate the convergence of TSAC.

\bibliographystyle{IEEEtran}
\bibliography{sample-base}




\end{document}